# Bidirectional Representation Learning from Transformers using Multimodal Electronic Health Record Data to Predict Depression

Yiwen Meng, William Speier, *Member*, Michael K. Ong and Corey W. Arnold, *Member, IEEE*

*Abstract*— **Advancements in machine learning algorithms have had a beneficial impact on representation learning, classification, and prediction models built using electronic health record (EHR) data. Effort has been put both on increasing models' overall performance as well as improving their interpretability, particularly regarding the decision-making process. In this study, we present a temporal deep learning model to perform bidirectional representation learning on EHR sequences with a transformer architecture to predict future diagnosis of depression. This model is able to aggregate five heterogenous and high-dimensional data sources from the EHR and process them in a temporal manner for chronic disease prediction at various prediction windows. We applied the current trend of pretraining and fine-tuning on EHR data to outperform the current state-of-the-art in chronic disease prediction, and to demonstrate the underlying relation between EHR codes in the sequence. The model generated the highest increases of precision-recall area under the curve (PRAUC) from 0.70 to 0.76 in depression prediction compared to the best baseline model. Furthermore, the self-attention weights in each sequence quantitatively demonstrated the inner relationship between various codes, which improved the model's interpretability. These results demonstrate the model's ability to utilize heterogeneous EHR data to predict depression while achieving high accuracy and interpretability, which may facilitate constructing clinical decision support systems in the future for chronic disease screening and early detection.**

## I. Introduction

Electronic health record (EHR) systems have become the main method of documenting patients' historical medical records over the last decade [1]. The latest report from the Office of the National Coordinator for Health Information Technology (ONC), stated that nearly 84% of hospitals have adopted at least a basic EHR system, a nine-fold increase from 2008 [2]. EHRs are composed of data from different modalities, documented in a sequence for each patient encounter, including demographic information, diagnoses, procedures, medications or prescriptions, clinical notes written by physicians, images, and laboratory results, which contribute to their high dimensionality and heterogeneity [3], [4]. Deep learning algorithms enable the usage of EHR data not only as a documenting method for billing purposes, but also as a source of tremendous amount of data to construct classification or prediction models, which build the foundation for creating clinical decision support systems and personalized precision medicine. However, there is an unsolved challenge of achieving high accuracy while providing adequate explanation for models' decision-making processes. Although several efforts have attempted to improve model interpretability [5]–[7], they did not address the problem of data heterogeneity that is pervasive in medical research as EHRs are often composed of data from various modalities in a sequential structure.

Depression is one typical comorbidity of chronic disease and a major cause of disability worldwide [8]. It often leads to a number of adverse outcomes, including increased risk of self-harm, premature mortality, and the development of comorbid general medical conditions, such as heart disease, stroke, and obesity [9]. Within a year of experiencing depression symptoms, patients are 4.4 times more likely to develop major depressive disorder (MDD) [10], a heterogeneous spectrum disorder with a variety of onsets, treatment responses, and comorbidities. The economic burden of individuals with MDD was $210.5 billion in 2010, which has increased from $173.2 billion in 2005 [11]. Although depression has become highly prevalent and costly, the current screening process used in clinics for patients with high risk of depression only produced a true positive rate of 50% [12]. Hence, much informatics effort has been put to increase the accuracy using machine learning algorithms [13]–[15].

The goal of this study is to create a model with high interpretability for predicting future diagnosis of depression while being able to accommodate the heterogeneity of EHR data and process it effectively in a temporal manner. We propose a Bidirectional Representation Learning model with a Transformer architecture on Multimodal EHR (BRLTM). This BRLTM is able to aggregate five EHR data modalities: diagnoses, procedure codes, medications, demographic

This work was supported by the National Heart, Lung, and Blood Institute (NIH/NHLBI R01HL141773).

Y. Meng, is with the Computational Diagnostics Lab, the Department of Bioengineering, at the University of California Los Angeles, 924 Westwood Blvd, Suite 420, CA 90024 USA (e-mail: lanyexiaosa@ucla.edu)

W. Speier is with the Computational Diagnostics Lab, the Department of Radiology at the University of California Los Angeles, 924 Westwood Blvd, Suite 420, CA 90024 USA (e-mail: speier@ucla.edu).

M.K. Ong is with the Department of Medicine at the University of California Los Angeles, 1100 Glendon Avenue, Suite 850, CA 90024 USA (e-mail:mong@mednet.ucla.edu).

C.W. Arnold is with the Computational Diagnostics Lab, the Department of Bioengineering, the Department of Radiology and the Department of Pathology at the University of California Los Angeles, 924 Westwood Blvd, Suite 420, CA 90024 USA (e-mail:cwarnold@ucla.edu)

information, and clinical notes. It enables modeling EHR data with the widely used two-stage pretraining and finetuning approach, which is significantly improved from previous works [4]–[6], [16]. The transformer architecture offers generalized representation learning on EHR data and the self-attention mechanism highly improves the model's interpretability by showing association of various EHR codes in sequences quantitatively [17]. Our results also improve the performance of bidirectional learning than forward-only methods in sequence modeling. This approach could help in clinical practice by identifying individuals potentially at risk for developing depression within a specific time interval who should be screened (and potentially treated) for depression.

## II. RELATED WORK

Performance on sequential or temporal learning tasks has been largely advanced by the advent of recurrent neural network (RNN) and its variations, including long-short-term-memory (LSTM) and gated recurrent unit (GRU). Previous studies have applied these methods on modeling time-series medical data, particularly on EHR data to predict future diagnoses [18], [19]. [5] first added a reverse time attention mechanism to RNN for heart failure prediction, which improved the model's interpretability by showing the temporal effect of events. [20] also achieved improving their model's interpretability using self-attention, but only applied it on diagnosis and procedure codes. [21] was able to predict clinical interventions from a deep neural network using lab results and demographics, but with a smaller number of features (34). MiME focused on learning the inner structure of an EHR by constructing a hierarchy of diagnosis level, visit level, and patient level embeddings [16]. The HCET model extended this hierarchical structure by removing the requirement of linked structure between diagnosis codes and procedure codes and medication while enabling attention on each EHR data modality to increase the model's interpretability [4]. However, these models did not provide a generalized approach to model EHR data with a two-stage pretraining and finetuning, nor reveal the latent association on every code instead of each aggregated modality in terms of model interpretability.

Recent developments in natural language processing (NLP) provide a number of potential methods applicable to EHR data. NLP uses sequential learning on word sequences, which can be applied to EHR data that is comprised of time series sequences from different data modalities. Dipole exhibited the potential of bidirectional learning on EHR data using an RNN with concatenated attention to predict diagnosis of diabetes [6] based on diagnosis and procedure codes. Choi et al. enhanced their model's latent representation learning with a graph convolution transformer (GCT), but did not perform sequential learning, focusing only on a single patient encounter [22]. The BEHRT model [23] first realized a two-stage transfer learning approach with the BERT model [24], but only applied it to diagnosis codes with a low dimensional feature space (301). Meanwhile, it merely relied on diagnosis codes as the true label for every disease, which actually reduced the prediction sensitivity or specificity, due to inaccuracy and incompleteness in International Classification of Disease (ICD) codes and they are mostly for billing purposes [25]. Thus, our BRLTM model aims to implement a generalized representation learning method on multimodal and high dimensional EHR data using the two-stage pretraining and finetuning approach, capable of aggregating five EHR modalities. In the meantime, we provided a feasible approach to increase the label accuracy for diagnosis of depression, which also enhanced the validity of model's prediction performance. Finally, this method also demonstrates the power of using incomplete or historical EHR sequences for future prediction whereas BERT model takes complete word sequences or sentences for downstream classification or text generation.

## III. DATA PREPROCESSING

We selected patients based on three primary diagnoses: myocardial infarction (MI), breast cancer, and liver cirrhosis, to capture a spectrum of clinical complexity. Generally, MI represents the least complexity, with acute onset, resolution, and straight-forward treatment. Breast cancer is increasingly complicated in terms of diagnoses and treatment options. Finally, a patient with liver cirrhosis may have many sequelae, generating a complex EHR representation. Patients for this work were identified from our EHR in accordance with an Institutional Review Board (IRB) (#14-000204) approved protocol. Each patient visit had EHR data types consisting of diagnosis codes in International Classification of Disease, ninth revision (ICD-9) format, procedure codes in Current Procedural Terminology (CPT) format, medication lists, demographic information, and clinical notes represented as 100 topics using latent Dirichlet allocation (LDA) analysis [26]–[28]. EHR data are structured in a tabular format in SQL, and transformed in a pandas data frame for Python coding [29]. All patient records coded with ICD-9 values for MI, breast cancer, or liver cirrhosis between 2006-2013 were included. Demographics were limited to the patient's gender and age. Initially, there were 45,208 patients and after preprocessing to remove visits without at least one ICD-9 code, CPT code, medication, or topic feature and eliminate patients with fewer than two visits, 43,967 patients were included in the analysis. More importantly, data after 15 days prior to the time of depression diagnosis was excluded for depressed patients to ensure no data leakage while all data were included for non-depressed ones. We followed the same data selection criteria as in the HCET model [4] with four prediction windows prior to the time of depression diagnosis: two weeks, three months, six months, and one year. The length of each data window was restricted to six months instead of patient's entire history to avoid bias towards patients with longer medical histories. For LDA feature generation, stop words were first removed from the note data. The MALLET [30] software package was used to fit the model with asymmetric priors. The model was fit over 1,000 iterations. Four example topics are shown in Table I.

TABLE I
EXAMPLE TOPICS

| Topic Num | Top five words |
|---|---|
| 55 | heart, cardiomyopathy, oht, failure, vt |
| 42 | liver, cirrhosis, ascites, lactulose, hepatic |
| 29 | value, component, creat, plt, hgb |
| 27 | transplant, tacrolimus, liver, renal, daily |

Patient Health Questionnaire (PHQ-9) scores [31], the most common way to identify the diagnosis of depression, were not available for this patient cohort during the data collection period. Hence, depression onset was identified by three methods: depression related ICD-9 codes, inclusion of an antidepressant drug in a patient's medication list, or appearance of an antidepressant drug in clinical notes, which has been used previously [4].

## IV. METHODS

In NLP implementations, BERT models process words sequentially. This method can be applied to EHR data by analyzing ICD-9 codes, CPT codes, medication lists, and topics as code sequences representing a patient's visits. Full ICD-9 codes are high dimensional that are sparsely represented with 9,285 distinct codes in our dataset. As in [19], dimensionality was reduced by grouping codes by the three digits before the decimal point to reduce its feature dimension to 1,131. Each demographic is added as an individual feature and repeated for every sequence. Pretraining was conducted through masked language modeling (MLM) to predict the mask code based on EHR sequences [24]. This pretraining is an unsupervised approach to learn the latent structure of EHR data as the result is able to show the association between two code in the sequence quantitatively. After pretraining, the saved model was added a classification head to finetune for the downstream task of chronic disease prediction. The feature dimension as the unique number of codes for five modalities were listed as follows:

- Diagnoses: 1,131
- Procedures: 7,048
- Medications: 4,181
- Demographics: 2
- Topics: 100

TABLE II
NOTATION USED IN THE FORMULATION OF BRLTM

| Notation | Definition |
|---|---|
| $V_t$ | EHR code sequences for $t^{th}$ visit, $t \in [1, L]$ |
| $L$ | Length of a patient's hospital visits |
| $X_i$ | One EHR code in patient's $t^{th}$ visit, $i \in [1, m_t]$ |
| $m_t$ | Number of codes in the $t^{th}$ visit |
| $D$ | Vocabulary of diagnosis codes |
| $C$ | Vocabulary of CPT codes |
| $M$ | Vocabulary of medications |
| $T$ | Vocabulary of topic features |
| $CLS$ | Starting point of one patient's EHR sequence |
| $SEP$ | A separation notation to separate codes in two consecutive visits |

### A. BRLTM model for EHR representation learning

Notation for the BRLTM model is included in Table II. Eq. (1) shows a patient's EHR, composed by different visits ranging from $V_1$ to $V_L$, where L is the length of the hospital visits. Two symbolic tokens $CLS$ and $SEP$ are adopted here: $CLS$ denotes the starting point of the EHR and $SEP$ denotes the separation between two consecutive visits. This formulates the input in Fig. 1.

$$EHR: (CLS, V_1, SEP, V_2, SEP, \ldots, V_L) \quad (1)$$

Each of the visit $V_t$ is comprised of EHR codes $X$, as shown in Eq. (2), where the number of codes is $m_t$, which varies for each visit. Every code is from the vocabulary of the dataset: $D$: diagnosis, $C$: procedure, $M$: medication and $T$: topics.

$$V_t: (X_1, X_2, \ldots, X_{m_t}), \quad X \in \{D, C, M, T\} \quad (2)$$

The original BERT models has three types of embeddings:

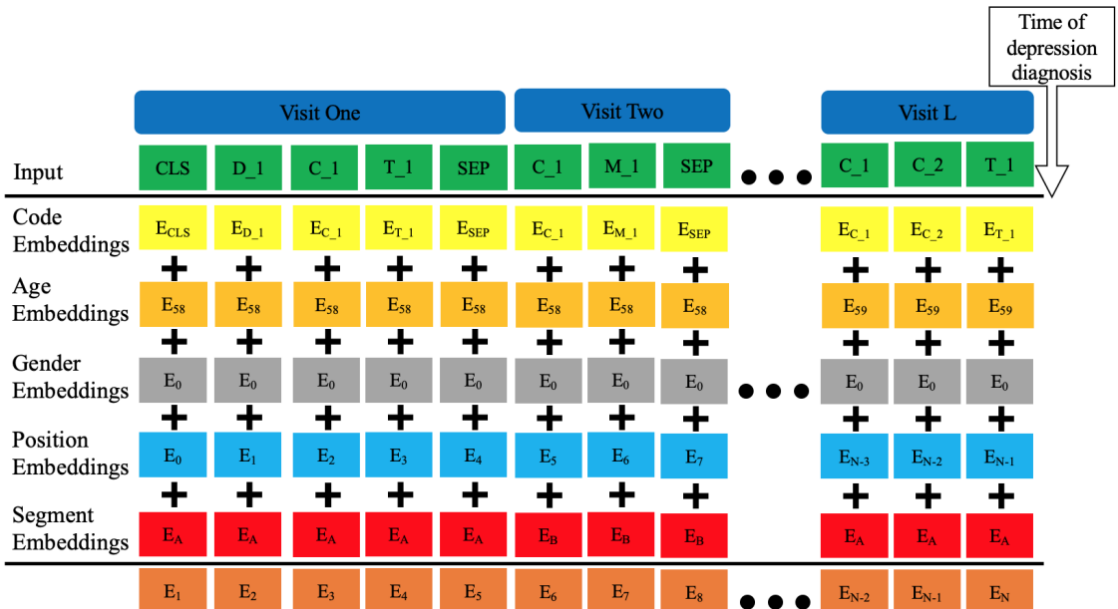

Fig. 1. Architecture of the BRLTM model for EHR representation learning. The subscripts show the original value for each embedding. CLS and SEP are symbolic tokens stands for the beginning of EHR and separation of two visits adjacent to each other, respectively. D, C, M, and T denote diagnoses, procedures, medications, and topics, respectively. The last row denotes the sum of the five embeddings as the output embedding.

token, position, and segment [24]. In our BRLTM model, we treated each token embedding as a code embedding and extended the model's ability to aggregate demographics by adding age and gender embeddings, shown in Fig. 1. There are five types of embeddings which are summed to generate the final output embeddings for training. Data after the depression diagnosis were excluded to avoid data leakage. Any data within 15 days prior to the diagnosis time was excluded to ensure the predictive power of the model. For non-depressed patients, the last time step of the EHR was substituted for the diagnosis time. Code embeddings are composed of four EHR data modalities mentioned in Eq. (2), while the original BERT model only takes word tokens as code embeddings. As in the original BERT model, position and segment embeddings indicate the position of one code in the full sequence which distinguishes codes in adjacent visits, which is highly efficient. Hence, we followed the same structure of them here and adopted pre-determined instead of learned encodings for positional embeddings to avoid weak learning of positional embedding due to high variety in a patient's sequence length. The position embeddings play an important role in sequence learning, equivalent to the recurrent structure in RNNs. Annotating the position of each code in the sequence enables the model to capture the positional interactions among EHR data modalities. However, position embeddings do not tell whether codes are from the same visit or not. Hence, segment embeddings are used to provide extra information to differentiate codes in adjacent visits by alternating between two trainable vectors, depicted as A and B in Fig. 1.

Age and gender embeddings are repeated in every position of the sequence. Combining code embeddings with the age embedding not only enables the model to use age information as a feature, but also provides temporal information in the sequence. As shown Fig. 2, the final embeddings are input into a bidirectional sequential learning step with transformer architecture as in the BERT model [24]. Hence, the latent contextual representation of five data modalities in temporal EHR sequences can be efficiently learned from the aggregation of these five embeddings. This architecture is capable of aggregating multimodal EHR data into a single model and processing them in a temporal manner, as well as investigating the inner association contingency between them in various visits. In total, the model can perform temporal representation learning from a patient's EHR. More importantly, it realizes the common two-stage transfer learning approach on EHR modeling, which has been widely adopted and has achieved outstanding performance in computer vision [32] as well as NLP [24], [33].

*B. Pretraining with Masked Language Modeling (MLM)*

An EHR is composed of multimodal code sequences, which is similar to the way that language is composed of word sequences. Hence, we hypothesized that the advantage of deep bidirectional sequential learning in language modeling over either a left-to-right model or the shallow concatenation of a left-to-right and a right-to-left model can be transferred to EHR modeling. As a consequence, we adopted the same pretraining approach of MLM from the original BERT paper [24]. Namely, we randomly selected 15% of EHR codes and modified them according to the following procedures:

- 80% of the time replace them with [MASK]
- 10% of the time replace them with a random EHR code
- 10% of the time do nothing and keep them unchanged

This structure in MLM forces the model to learn the distributional contextual representation between EHR codes as the model does not know which codes are masked or which codes have replaced by a random code. EHR modeling is not affected significantly because only 1.5% (10% of 15%) of codes are randomly replaced. This random replacement brings a small perturbation that distracts the model from learning the true contextual sequences of the EHR and forces the model to identify the noise and continue learning the overall temporal progression. We followed the precision score (true positives divided by predicted positives) at a threshold of 0.5 as the metric to evaluate pre-training MLM task, which was same approach as the BERHT model [23]. The average value is calculated over every masked code and over all patients. We followed results from previous models [23], [24] with random search to find the best set of hyperparameters during training. In addition, we conducted an ablation study to investigate the contribution of each data modality by training the model using all five modalities first and training again after removing topics and CPT codes individually while training without both of them as the last comparison. The data used for MLM is shown in the second column of Table III, displaying the total number of patients for each of three primary diagnoses, and the number of patients depressed patient displayed in the brackets. Note that some patients had more than one primary diagnosis. Furthermore, distribution of age, gender and EHR sequence length are also displayed here.

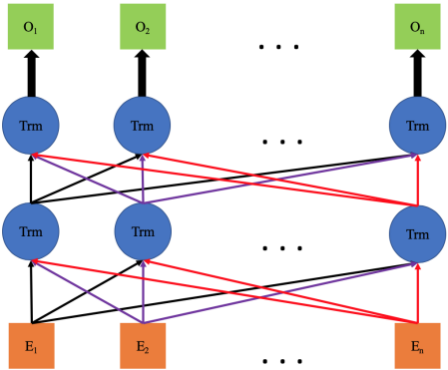

Fig. 2. Illustration of bidirectional learning with transformer architecture. The orange squares are final embeddings in Fig. 1, which are the input sequences here. Trm stands for the transformer while the green squares denote the output sequence. O denotes the output for each code after learning.

TABLE III
STATISTICS OF DATASETS FOR TWO TRAINING APPROACHES

| Datasets | pretraining | finetuning |
|---|---|---|
| Patients with MI | 10,616 (2,915 depressed) | 2,943 (1,280 depressed) |
| Patients with breast cancer | 23,3077 (4,483 depressed) | 5,568 (1,960 depressed) |
| Patients with liver cirrhosis | 11,757 (2,359 depressed) | 2,218 (772 depressed) |
| Gender | 70.18% female | 72.54% female |
| Age | 65.78 ± 14.99, min: 18, max 100 | 68.78 ± 15.46, min: 18, max 98 |
| Sequence length | 54.64 ± 45.37, min:2, max: 1,186 | 54.64 ± 45.37, min:2, max: 180 |

### C. Fine tuning to predict diagnosis of depression

After pretraining to learn the latent contextual representation of the EHR, one feed-forward classification layer was added for finetuning on a specific dataset. We followed the same data selection criteria as in the HCET model [4] with four prediction windows prior to time of depression diagnosis: two weeks, three months, six months, and one year. The length of each data window was restricted to six months instead of patient's entire history to avoid bias towards patients with longer medical histories. Patients who had at least one ICD-9, CPT, medication or topic feature in all four time windows were included. After processing data based on this method, 10,148 patients were selected, where 3,747 were diagnosed with depression. Basic statistics of the data for this prediction task are shown in the third column of Table III. Finally, predicting performance of each model was evaluated in receiver characteristic area under curve (ROCAUC) and PRAUC.

### D. Training Details

The BERT model was implemented in Pytorch 1.4 and trained on a workstation equipped with an Intel Xeon E3-1245, 32 GB RAM and a 12G NVIDIA TitanX GPU. We followed the training scheduler with the Adam [34] optimizer used the original BERT model [24] and set the warmup proportion and weight to 0.01 and 0.1, respectively. The Gaussian error linear unit (GELU) rather than the standard ReLu was used as the non-linear activation function in the hidden layers. Pretraining of MLM used the first dataset with the minibatch of 256 patients for 100 epochs and evaluated at every 20 iterations. The dataset for Finetuning of the prediction task underwent 10 random data splits: 70% training, 10% validation, and 20% test, and trained with minibatch of 64 patients for 50 epochs. Dropout of 0.1 was set to both hidden size and multi-head attention to address overfitting. The source code and more detailed description of the model is available at https://github.com/lanyexiaosa/brltm.

### E. Baseline models

The following models described in Section II were used to compare the prediction performance to our BRLTM model: Dipole [6], MiME* [16], HCET [4], BERHT [23]. We modified MiME to MiME* as the original MiME model requires external knowledge of linked relation between ICD-9 codes and associated CPT codes and medication lists during each visit, which was not applicable to this dataset. Paired t-test was used to compute the statistical significance when

TABLE IV
RESULTS OF PRETRAINING WITH MLM

| Data combination | All | No topic | No CPT | No (topic+CPT) |
|---|---|---|---|---|
| Vocabulary size | 12,460 | 12,360 | 5,412 | 5,312 |
| Precision | 0.4248 | 0.4324 | 0.4836 | 0.5086 |
| Learning rate | 1e-4 | 1e-4 | 1e-4 | 1e-4 |
| Embedding size | 216 | 240 | 252 | 264 |
| Attention layers | 9 | 9 | 6 | 6 |
| Attention heads | 12 | 12 | 12 | 12 |
| Intermediate layer | 512 | 512 | 256 | 256 |

All means using all five modalities. The result of MLM in BEHRT model as the last column is compared here.

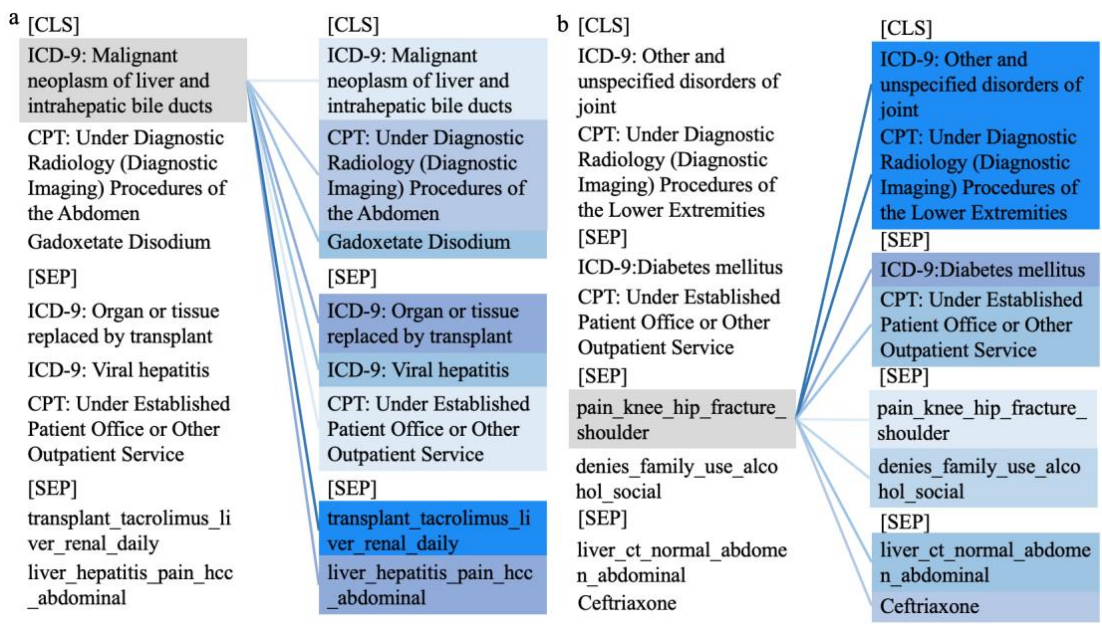

Fig. 3. Quantitative analysis of self-attention from two patients' EHR sequences shown in color plots. CLS and SEP represent the beginning of the record and separators between visits, respectively. Topic features are represented as the five most commonly associated words. Each example is presented as two identical columns as the left one represents the code of interest colored in grey while the right one indicates the corresponding associations to the highlighted code on the left. The intensity of the blue color on the right column denotes the strength of the attention score; the deeper blue color suggests higher self-attention score and hence the stronger the latent association.

comparing two models.

## V. RESULTS

### A. Pretraining with MLM and self-attention of code sequences

Table IV presents the results for MLM including the optimal hyperparameter settings on various combination of EHR data modalities. According to the result, the precision score raises gradually from 0.4208 to 0.5086 as the vocabulary size decreases by excluding more data modalities. The optimal embedding size also follows this trend as 216 for all data and 264 for data without topics and CPT. The number of attention layers and the number of multi-head has the opposite trend, changing from 9 to 6 and 512 to 256, respectively.

After training, attention weights for any EHR sequence can be extracted by enabling the output variable in the BERT encoder function and better presented via this visualization tool: https://github.com/jessevig/bertviz. Fig. 3 (a) and (b) exhibits the self-attention weights from two patients' EHR sequences, retrieved from the attention component of the last layer of the BERT model. It was illustrated in Fig. 3(a) that this patient was diagnosed with malignant neoplasm of the liver. The self-attention weight indicated its highest association with the topic feature associated with the words "transplant, tacrolimus, liver, renal, and daily" and second highest relation to the ICD-9 code for "Organ or tissue replaced by transplant." The topic feature with the words "liver, hepatitis, pain, hcc, and abdominal" described the fact that the patient was undergoing a liver transplant after the original diagnosis. Fig. 3(b) displays an example patient diagnosed initially with an unspecified joint disorder which led to a topic feature of "pain, knee, hip, fracture, and shoulder" shown later in the EHR sequence. The darker color suggests the stronger association of this topic feature to the original diagnosis code (diagnostic radiology imaging) and a weaker latent relation to the diagnosis of diabetes and the medication ceftriaxone. The attention scores demonstrate the association of the patient's health status with an original diagnosis of joint disorder and a comorbidity of diabetes developed later. This matches the meta-analysis that arthritic patients have 61% higher odds of having diabetes compared to the population without arthritis [35].

### B. Comparison of Performance in Depression Prediction

Table V shows the ROCAUC and PRAUC from all baseline models and our BRLTM model at the four prediction windows. The BRLTM model achieved the highest performance in each prediction window with statistically significant improvements over the next best model (HCET). BEHRT generated slightly better results than MiME* in the two shortest time windows, but MiME* reached higher numbers in longer windows. This result follows those observed in HCET where ICD-9 possessed attention weights higher than the average in smaller prediction windows while it was lower than the average in larger windows. The prediction performance was slightly improved with Dipole which used ICD-9 and CPT codes in a bidirectional learning method. Finally, there is a consistent decrease of accuracy for every model as the prediction window moves further away from the time of diagnosis.

### C. Prediction performance for each primary diagnosis

Table VI displays the individual results for each of three primary diagnoses in prediction windows of two-weeks and one-year. Our BRLTM model also achieved the best performance for all three primary diagnoses in both prediction windows. The ROCAUC from all three diseases within each model is similar even though the number of patients with breast cancer was substantially higher than the other diseases (n=5,568), which indicates no bias toward any primary diagnosis for the task of predicting diagnosis of depression. It is notable that while the PRAUC for patients with myocardial infarction is relatively higher than other two, the difference in the BRLTM model is relatively small. Dipole generated a mean increase of around 0.02 both in ROCAUC and PRAUC

TABLE V
COMPARISON OF PREDICTION PERFORMANCE FOR DIFFERENT MODELS

| Prediction window | Two weeks | | Three months | | Six months | | One year | |
|---|---|---|---|---|---|---|---|---|
| Models | ROCAUC | PRAUC | ROCAUC | PRAUC | ROCAUC | PRAUC | ROCAUC | PRAUC |
| MiME* ICD-9+CPT+Medication | 0.76 (0.01) | 0.67 (0.02) | 0.74 (0.01) | 0.64 (0.02) | 0.72 (0.02) | 0.61 (0.01) | 0.70 (0.01) | 0.61 (0.01) |
| BEHRT ICD-9 | 0.77 (0.02) | 0.68 (0.01) | 0.75 (0.02) | 0.65 (0.01) | 0.71 (0.01) | 0.61 (0.02) | 0.69 (0.02) | 0.60 (0.02) |
| Dipole ICD-9+CPT | 0.78 (0.02) | 0.70 (0.01) | 0.76 (0.02) | 0.67 (0.02) | 0.75 (0.01) | 0.65 (0.02) | 0.74 (0.01) | 0.64 (0.01) |
| HCET All | 0.81 (0.01) | 0.73 (0.01) | 0.80 (0.01) | 0.70 (0.02) | 0.79 (0.01) | 0.69 (0.01) | 0.78 (0.01) | 0.67 (0.01) |
| BRLTM All | **0.85 †** **(0.02)** | **0.78 †** **(0.01)** | **0.84 †** **(0.01)** | **0.76 †** **(0.01)** | **0.83 †** **(0.01)** | **0.74 †** **(0.02)** | **0.81 †** **(0.01)** | **0.73 †** **(0.01)** |

Values in parenthesis stand for standard deviations across randomizations and bold values denotes the highest in each column. † indicates the value is significantly better than that from the best baseline model HCET ($p<0.05$). The words after each models denotes the input data modalities where all means all five in our dataset.

TABLE VI
COMPARISON OF PREDICTION PERFORMANCE FOR THREE PRIMARY DIAGNOSES

| Prediction window | Two weeks | | | | | | One year | | | | | |
|---|---|---|---|---|---|---|---|---|---|---|---|---|
| Diseases | Breast cancer | | MI | | Liver cirrhosis | | Breast cancer | | MI | | Liver cirrhosis | |
| Models | ROC AUC | PR AUC | ROC AUC | PR AUC | ROC AUC | PR AUC | ROC AUC | PR AUC | ROC AUC | PR AUC | ROC AUC | PR AUC |
| MiME* | 0.77 (0.01) | 0.67 (0.02) | 0.75 (0.01) | 0.70 (0.02) | 0.76 (0.02) | 0.67 (0.01) | 0.71 (0.02) | 0.61 (0.01) | 0.69 (0.02) | 0.64 (0.01) | 0.70 (0.01) | 0.61 (0.02) |
| BERHT | 0.78 (0.02) | 0.68 (0.01) | 0.77 (0.01) | 0.71 (0.02) | 0.77 (0.01) | 0.68 (0.02) | 0.70 (0.01) | 0.59 (0.02) | 0.70 (0.01) | 0.62 (0.01) | 0.69 (0.02) | 0.60 (0.02) |
| Dipole | 0.79 (0.02) | 0.69 (0.02) | 0.78 (0.01) | 0.71 (0.02) | 0.78 (0.02) | 0.69 (0.01) | 0.75 (0.01) | 0.63 (0.02) | 0.74 (0.02) | 0.66 (0.02) | 0.74 (0.01) | 0.63 (0.01) |
| HCET | 0.81 (0.01) | 0.73 (0.01) | 0.79 (0.01) | 0.77 (0.01) | 0.80 (0.01) | 0.72 (0.01) | 0.78 (0.01) | 0.67 (0.01) | 0.77 (0.01) | 0.71 (0.01) | 0.77 (0.01) | 0.66 (0.01) |
| **BRLTM** | **0.85 (0.01)** | **0.76 (0.01)** | **0.85 (0.01)** | **0.78 (0.01)** | **0.84 (0.01)** | **0.75 (0.01)** | **0.80 (0.01)** | **0.72 (0.01)** | **0.81 (0.01)** | **0.74 (0.01)** | **0.80 (0.01)** | **0.71 (0.01)** |

across all diseases. In addition, HCET achieved better values than Dipole with higher improvement in PRAUC than ROCAUC. The BRLTM model further improved the performance from HCET with highest increase of 0.06 in ROCAUC for MI in the window of two weeks and 0.05 in PRAUC for breast cancer and liver cirrhosis in the one-year window.

Fig. 4 contains the confusion matrices individually for three primary diagnoses in the two-week prediction window from four models. The output probability was calibrated using isotonic regression [36] using a threshold of 0.5, and numbers were aggregated from 10-fold cross validation. The class distribution was imbalanced with a smaller portion of depressed patients for each primary diagnosis. MiME* reached a higher portion of false positive than false negative and Dipole managed to reduce both numbers slightly. Our BRLTM model significantly decreased the false positives by almost 50% from HCET while reducing false negatives by roughly 40% for MI and 30% for breast cancer and liver cirrhosis. Hence, it achieved outstanding average precision and recall of 0.94 and 0.84, respectively, over the three primary diagnoses.

|  | Breast cancer | | Liver cirrhosis | | MI | |
|---|---|---|---|---|---|---|
| BEHRT | Actual 0 | 3329 / 279 | Actual 0 | 1321 / 125 | Actual 0 | 1468 / 195 |
| | 1 | 877 / 1173 | 1 | 267 / 505 | 1 | 534 / 746 |
| | | 0  1 Predicted | | 0  1 Predicted | | 0  1 Predicted |
| Dipole | Actual 0 | 3357 / 251 | Actual 0 | 1338 / 108 | Actual 0 | 1479 / 184 |
| | 1 | 694 / 1266 | 1 | 207 / 565 | 1 | 460 / 820 |
| | | 0  1 Predicted | | 0  1 Predicted | | 0  1 Predicted |
| HCET | Actual 0 | 3442 / 166 | Actual 0 | 1362 / 84 | Actual 0 | 1524 / 139 |
| | 1 | 552 / 1408 | 1 | 153 / 619 | 1 | 294 / 986 |
| | | 0  1 Predicted | | 0  1 Predicted | | 0  1 Predicted |
| BRLTM | Actual 0 | 3528 / 80 | Actual 0 | 1397 / 49 | Actual 0 | 1592 / 71 |
| | 1 | 378 / 1582 | 1 | 109 / 663 | 1 | 174 / 1106 |
| | | 0  1 Predicted | | 0  1 Predicted | | 0  1 Predicted |

Fig. 4 Confusion matrices for patients separated by three primary diagnosis at a window of two weeks for four models. The numbers are aggregated together with 10-fold cross validation. Label 0 means non-depressed while 1 means depressed.

The computation time for every model is shown in Table VII, where values were averaged over batches for training and testing. According to the result, BRLTM obtained the highest computation time both in training and test over baseline models, while MiME* reached the lowest.

TABLE VII
COMPUTATION TIME FOR EACH MODEL PER BATCH

| Models | Training (s) | Testing (s) |
|---|---|---|
| MIME* | 0.78 | 0.57 |
| BEHRT | 1.55 | 1.03 |
| Dipole | 1.23 | 0.89 |
| HCET | 0.94 | 0.67 |
| BRLTM | 2.37 | 1.83 |

VI. DISCUSSION

According to results in Tables V and VI, our BRLTM model sufficiently resolved the data heterogeneity issue by realizing bidirectional sequential learning and enabling the stucture to aggregate multimoal EHR data, which achieved the best performance in predicting future diagnoisis of depression in four prediction windows. Additionally, the comparison to other models demonstrates the advantage of including more data modalities for predictive power as the BEHRT model only took diagnosis codes in their study. On the other hand, the better results from Diople than MiME* validates the advantage of bidirectional learning over single direction, as medication was less frequently present than ICD-9 and CPT, which Dipole did not take as the input. However, its lower performance than HCET, which adopted the forward-only sequnetial learning, highlights the importance to aggregate topics feature and demographics as Dipole only input ICD-9 and CPT codes, while HCET was capable of including all five modalities. Meanwhile, these results indicate that each model's performance consistently declines as the prediction window moves further away from the diagnosis time point, which agrees with our expectation that records closer to the diagnosis are more likely to contain relevant information and provide better predictions. Table IV shows the observation that for a larger vocabulary size or more data modalities, a smaller embedding size should be used, but the number of attention layers and intermediate layer size should be increased, while no strong perference of the learning rate and number of attention heads. The results also approves model's flexible structure of several tunable hyperparameters, espeically in attetnion layers, enabling it to process various types of EHR data which may be collected from different insitutions.

The BRLTM model demonstrated another advantage of improving model's interpretability by quantitalively revealing the latent association between codes in the sequence using self-attetion and multi-head attetion. More importantly, we sucessfully realized the common two-stage transfer learning apporach of pretraining and finetuing on modeling EHR data. Privacy issues related to EHR data, restrict the ability of institutions to share data, which substantially hinders the development in this field. The two-stage approach allows institutions with access to large amount of EHR data to provide the pretrained model as a general EHR feature extractor so that others can take the advantage by only finetuning the pretrained model on the customized dataset for specific tasks [37], [38]. This process benefits EHR representation learning and lays the foundation to allow adequate predictive power for models built on small EHR datasets. Furthermore, the BRLTM model also provides a generalized architecture that can be adopted with every EHR system by increasing the vocabulary of code embedding or by stacking more embedding layers for addition data modalities not used in this work.

The results from our BRLTM model and the previous HCET model both emphasize the critical contribution of clinical notes to build predictive models from EHR. We used topic modeling with LDA to extract semantic features of topics, which was limited by the bag of words assumption. Thus, future studies could utilize more recent NLP tools, such as BERT [24] or GPT-2 [33] to optimize contextual representation of clinical notes, which could further improve the overall performance. In addition, our model's predictive power was exhibited in predicting future diagnosis of depression, which was a binary classification task. Future studies could expand model's robustness in prediction by performing multiclass prediction simultaneously for other chronic diseases, such as hypertension, diabetes, and obesity. Reducing labeling errors is also critical to build an accurate machine learning model. Only relying on diagnosis codes as the label of every disease injects noise in the label, which was widely adopted in previous studies [16], [23]. Instead, we adopted three criteria for determining depression diagnosis, which mitigated the labeling error compared to previous studies. Future prospective studies could continue the effort to acquire more precise labels for EHR, such as administering PHQ-9 surveys periodically to increase precision in depression diagnosis. Thus, more robust predictive models could be constructed to track the disease progression as well as early detection, which could provide more applications for the BRLTM model in future work. Finally, all the results were based on this patient cohort and EHR data, hence conducting more testing on EHR data collected from various cohorts as wells as predicting different chronic diseases could bring more insights on the generalizability of BRLTM model. In addition, other EHR data modalities such as laboratory results [21] were not included in the BRLTM model as they were unavailable for collection. Future studies may extend the model to aggregate more data modalities to further make it as a generalized model.

## VII. CONCLUSION

We have developed a bidirectional deep learning model BRLTM to perform temporal representation learning on multimodal EHR data and successfully realized two-stage pretraining and finetuning. These efforts contributed to significant improvement on chronic disease prediction as well as advancement of model interpretability by the quantitative analysis of self-attention weights of EHR sequences. The results demonstrate the ability of the two-stage transfer learning approach for EHR modeling to overcome limitations in the amount of available data and that bidirectional learning can provide superior performance to unidirectional. This approach facilitates the development of clinical decision support systems for chronic disease prediction, such as a screening tool for patients at high risk of depression, thereby enabling early intervention. Future works could test the model with more EHR data modalities, increasing the labeling precision of depression diagnosis or improving semantic feature extraction on clinical notes.